\documentclass[10pt,twocolumn,letterpaper]{article}
\usepackage{tikz}
\usepackage{iccv}
\usepackage{times}
\usepackage{epsfig}
\usepackage{graphicx}
\usepackage{amsmath}
\usepackage{amssymb}
\usepackage{multirow}
\usepackage{float}
 
\usepackage[pagebackref=true,breaklinks=true,letterpaper=true,colorlinks,bookmarks=false]{hyperref}

\usepackage[accsupp]{axessibility}
\usepackage{commath}
\usepackage{fancyhdr}
\usepackage{yhmath}
\usepackage[ruled,vlined]{algorithm2e}
\usepackage{colortbl}
\definecolor{Gray}{gray}{0.9}
\usepackage{cuted}
\usepackage{multicol}
\usepackage{ragged2e}
\iccvfinalcopy 

\def\iccvPaperID{3497} 

\newcommand\blfootnote[1]{%
  \begingroup
  \renewcommand\thefootnote{}\footnote{#1}%
  \addtocounter{footnote}{-1}%
  \endgroup
}

\ificcvfinal\fi

\begin{document}

\title{Domain Adaptive Few-Shot Open-Set Learning}

\author{Debabrata Pal\textsuperscript{1}, Deeptej More\textsuperscript{2,\#}, Sai Bhargav\textsuperscript{1,\#}, Dipesh Tamboli\textsuperscript{3}, Vaneet Aggarwal\textsuperscript{3}, Biplab Banerjee\textsuperscript{1}\\
\textsuperscript{1}Indian Institute of Technology, Bombay, \textsuperscript{2}Manipal Institute of Technology, India,\textsuperscript{3}Purdue University\\
{\tt\small debabrata.pal@iitb.ac.in, deeptejrane16@gmail.com, sai.bhargav@iitb.ac.in,}\\
{\tt\small dtamboli@purdue.edu, vaneet@purdue.edu, getbiplab@gmail.com}
}

\twocolumn[{
\maketitle
\centering
\vspace{-9mm}
  \includegraphics[width = \linewidth, height=3cm]{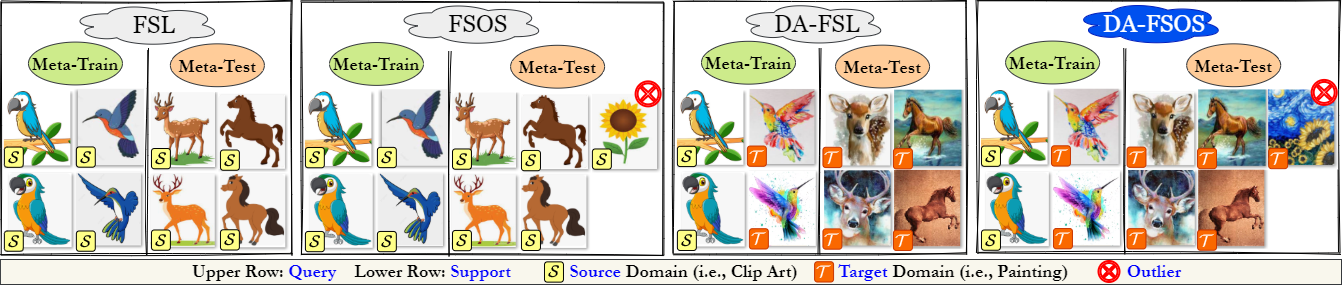} 
  \\
   \hspace{-0.2mm}\small \justifying {Figure 1. Domain Adaptive Few-Shot Open-Set Learning (DA-FSOS) tackles the challenges of both Domain Adaptive Few-Shot Learning (DA-FSL) and Few-Shot Open-Set Learning (FSOS) by integrating them into a unified framework. In training phase, the model observes a fully-supervised source domain $\mathcal{S}$ and a sparsely-supervised target domain $\mathcal{T}$, where the labels of both domains are disjoint. However, during testing, the model encounters a few-shot support set of new \textit{known} classes from $\mathcal{T}$, while the test query set contains unlabeled samples from both the known and previously unknown classes. \textbf{{Project page:}} \textcolor{magenta}{\url{https://github.com/DebabrataPal7/DAFOSNET}} }
   \vspace{0.15in}
   \label{fig:Teaser1}
}]  

\maketitle
\ificcvfinal\thispagestyle{empty}\fi

\section*{\centering Abstract}
\vspace{-0.1in}
\textit{Few-shot learning has made impressive strides in addressing the crucial challenges of recognizing unknown samples from novel classes in target query sets and managing visual shifts between domains. However, existing techniques fall short when it comes to identifying target outliers under domain shifts by learning to reject pseudo-outliers from the source domain, resulting in an incomplete solution to both problems.
To address these challenges comprehensively, we propose a novel approach called Domain Adaptive Few-Shot Open Set Recognition (DA-FSOS) and introduce a meta-learning-based architecture named \textsc{DAFOS-Net}. During training, our model learns a shared and discriminative embedding space while creating a pseudo-open-space decision boundary, given a fully-supervised source domain and a label-disjoint few-shot target domain.
To enhance data density, we use a pair of conditional adversarial networks with tunable noise variances to augment both domains' closed and pseudo-open spaces. Furthermore, we propose a domain-specific batch-normalized class prototypes alignment strategy to align both domains globally while ensuring class-discriminativeness through novel metric objectives.
Our training approach ensures that \textsc{DAFOS-Net} can generalize well to new scenarios in the target domain. We present three benchmarks for DA-FSOS based on the Office-Home, mini-ImageNet/CUB, and DomainNet datasets and demonstrate the efficacy of DAFOS-Net through extensive experimentation. }\blfootnote{\textsuperscript{\#}Deeptej and Sai have contributed equally to this work.}

\section{Introduction}
The development of deep learning techniques has led to significant advancements in visual recognition tasks by leveraging their ability to learn data-driven features from a vast amount of training examples \cite{resnet, alexnet}. However, labeling is a laborious and costly task, which is why few-shot learning (FSL) \cite{protonet,lee2019meta} aims to recognize target classes with limited supervision by learning transferable features from a label-disjoint source domain that has sufficient supervision.

Despite progress \cite{wang2020generalizing}, FSL models encounter difficulties in two critical scenarios: \textbf{i}) when the under-represented target domain classes are drawn from a different distribution than the source domain. One solution to this is \textit{Domain Adaptive Few-Shot Learning} (DA-FSL), which adapts disjoint classes from the fully-supervised source and sparsely-supervised target domains during training \cite{zhao2021domain, cdfsl1, cdfsl2}, and  \textbf{ii}) \textit{Few-Shot Open-Set Learning} (FSOS) \cite{liu2020few}, which assumes no domain gap between source and target but combines FSL with \textit{Open-Set Recognition} (OSR) \cite{osrsurvey,bendale2016towards} to detect novel class samples as outliers during testing. However, it is desirable for a generic FSL system to handle domain discrepancy and reject outliers during inference in a unified manner for practical applications. To this end, we introduce a novel scenario, called \textit{Domain Adaptive Few-Shot Open-Set Learning} (DA-FSOS), which is illustrated in Fig \textcolor{red}{1}.  

DA-FSOS is a method that addresses a challenging problem: \textit{how to learn from supervision in one domain while adapting to a different domain with non-overlapping known classes and potential test-time outliers under a few-shot setting}. During training, the method uses supervision from the source domain and a few-shot training set from a disjoint set of classes of the target domain. However, during testing, the method is expected to handle unlabelled samples from a new set of \textit{known} and \textit{open-set} classes from the target domain, adding an extra challenge. DA-FSOS can be applied to self-driving cars. By learning to identify known classes and reject outliers from abundant gaming data, DA-FSOS can help detect real-world objects of interest (known and unknown) on the road.

Also, to detect novel animal species, unknown land cover objects, unknown viruses in medical imaging, etc., DA-FSOS can become instrumental.

DA-FSOS combines several techniques to tackle these challenges, including domain adaptation, few-shot learning, and open-set recognition. The setting is designed to handle unconstrained domain differences, extremely limited supervision in the target domain, and the absence of prior knowledge regarding the target open space in test time.

While DA-FSL and FSOS techniques can be combined, they may not be sufficient to effectively solve the DA-FSOS problem as they rely on different assumptions individually (see Section \ref{section:exp}). Similarly, combining an FSL technique with open-set DA \cite{osda} or a DA-FSL model with the OSR approach may not be suitable since DA variants assume label space consistency between domains, which is not the case for DA-FSOS. \textit{Therefore, it is necessary to develop a model specifically designed to solve DA-FSOS}.

\textbf{Our proposed \textsc{DAFOS-Net}:} 
In this paper, we present a novel model called \textsc{DAFOS-Net} that addresses the DA-FSOS problem by integrating three crucial considerations. Our approach aims to learn a prototype-based classifier that can reject target outliers during testing, and we propose a meta-training method that simulates the test scenario by dividing the available training classes from both domains into known and pseudo-unknown categories in each episode.

To address the issue of overfitting caused by limited target supervision, we propose a \textbf{\textit{data augmentation}} technique in \textsc{DAFOS-Net}. Unlike previous FSOS methods that only augment known classes from the source domain \cite{pal2023morgan}, our approach involves augmenting both known and pseudo-unknown classes from both source and target domains using a pair of class and domain conditional adversarial networks (cGANs). To maintain feature consistency when generating known samples, we apply a low noise variance for the respective cGAN. However, to increase the scatter when generating pseudo-unknown samples, we use a high noise variance. To prevent mode collapse during training, we introduce a regularizer that ensures the outputs of the two cGANs are consistently different.

We also introduce \textbf{\textit{Global Cross-Domain Prototype Alignment (GCDPA)}} strategy by exploiting domain-specific batch-norm statistics to bring the domains closer for the smooth transfer of source knowledge into the target. 

However, we must ensure that the adaptation does not cause misalignments between the domains. To maintain class discriminativeness, we introduce a class compactness loss for the known class samples. Simultaneously, a novel {\textbf{\textit{prototype diversification loss}} objective is introduced to maximize the gap between the known-class prototypes and a combination of in-domain pseudo-unknown samples with the across-domain class prototypes.

Finally, we propose a learning-based approach to automatically predict the domain and inlier/outlier class labels for the test samples. This approach eliminates the need for manual threshold selection \cite{liu2020few} and improves the ability to solve \textbf{\textit{generalized DA-FSOS}} by selectively considering either source or target prototypes based on the predicted domain labels for classifying the test queries.

In summary, our \textbf{novel contributions} are as follows:

\noindent 1. We introduce DA-FSOS, a practical problem setting that generalizes FSOS and DA-FSL, and to solve that, we propose an end-to-end solution called \textsc{DAFOS-Net}.

\noindent 2. \textsc{DAFOS-Net} integrates four novel ideas:
\textbf{i}) An episodic training strategy to develop a domain-agnostic open-set classifier.
\textbf{ii}) A generative feature augmentation scheme that produces diversified known and pseudo-unknown class samples for both domains from few-shot training samples. 
\textbf{iii}) A metric objective that ensures class compactness and discriminativeness for both domains. 
\textbf{iv}) A prototypical batch-norm alignment-based global domain adaptation.

\noindent 3. We present the \textit{standard} and \textit{generalized} DA-FSOS training and evaluation protocols. Accordingly, we evaluate our approach on DomainNet\cite{domainnet}, miniImageNet-CUB \cite{cub, matchingnet}, and Office-Home \cite{offhome} datasets and achieve an average gain of 15\% closed accuracy and 17\% AUROC.

\renewcommand{\thefigure}{2}
\begin{figure*}[!htbp]
    \centering
    \includegraphics[width=\linewidth]{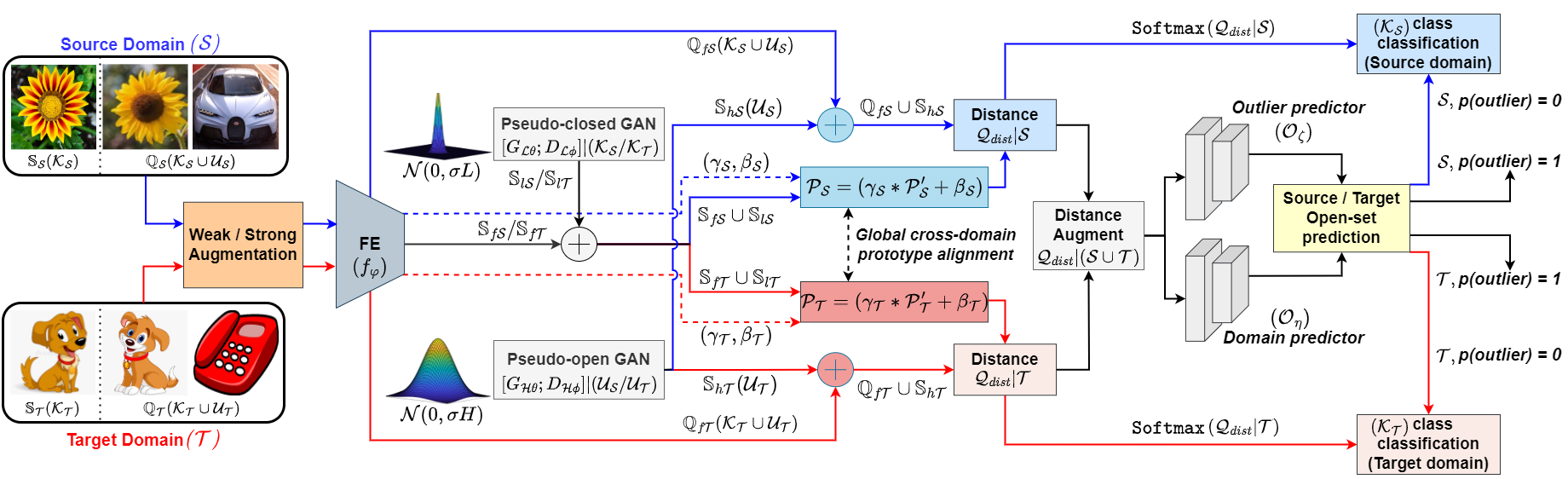}
    \vspace*{-1mm}
    \caption{The model architecture of \textsc{DAFOS-Net}. In each episodic training, Support $\mathbb{S}_{\mathcal{S}}(\mathcal{K}_{\mathcal{S}}) / \mathbb{S}_{\mathcal{T}}(\mathcal{K}_{\mathcal{T}})$ and Query samples $\mathbb{Q}_{\mathcal{S}}(\mathcal{K}_{\mathcal{S}}\cup \mathcal{U}_{\mathcal{S}}) / \mathbb{Q}_{\mathcal{T}}(\mathcal{K}_{\mathcal{T}}\cup \mathcal{U}_{\mathcal{T}})$ from $\mathcal{D}_s$ and $\mathcal{D}_d$ with $\mathcal{K}_{\mathcal{S}}$/$\mathcal{K}_{\mathcal{T}}$ known and $\mathcal{U}_{\mathcal{S}}$/$\mathcal{U}_{\mathcal{T}}$ pseudo-unknown classes are considered from $\mathcal{S}$ and $\mathcal{T}$. The domain and class conditional GANs ($G_{\mathcal{L}\theta}, D_{\mathcal{L}\phi}$) and $(G_{\mathcal{H}\theta}, D_{\mathcal{H}\phi})$ separately synthesize features for the known and pseudo-unknown classes. The augmented support sets $\mathbb{S}_{f \mathcal{S}} \cup \mathbb{S}_{l \mathcal{S}}$ and $\mathbb{S}_{f \mathcal{T}} \cup \mathbb{S}_{l \mathcal{T}}$ are used to produce the class prototype embeddings $\mathcal{P}'_{\mathcal{S}}$ and $\mathcal{P}'_{\mathcal{T}}$ for both the domains, which are aligned through the learnable domain-specific batch-norm parameters $(\gamma_{\mathcal{S}}, \beta_{\mathcal{S}})$ and $(\gamma_{\mathcal{T}}, \beta_{\mathcal{T}})$. Finally, a pair of domain-agnostic classifiers ($\mathcal{O}_{\zeta}, \mathcal{O}_{\eta}$) are deployed for the purpose of domain and inlier/outlier class predictions for the query samples.}
    \label{fig:Architecture}
    \vspace*{-1mm}
\end{figure*}

\section{Related works}

\noindent \textbf{Few-shot learning:} 
Given the significance of FSL, various techniques have been introduced in computer vision and natural language processing. 
Among them, meta-learning has received substantial attention, owing to the advantages of the episodic training strategy in facilitating \textit{learning to learn}. The metric learning-based meta-learning \cite{protonet, matchingnet, relationalnet} focus on distance metric learning for nearest neighbour (NN) search, aiming to learn a discriminative embedding space from low-shot support samples. Model-based FSL \cite{mlssl, finn2017model} leverage novel optimization algorithms, rather than gradient descent, to fit in the few-shot regime. Additionally, several FSL models rely on data augmentation \cite{fslaug1, fslaug2} and self-supervision \cite{ssl1, ssl2} to improve performance. For a comprehensive survey of FSL, see \cite{wang2020generalizing, fsls2, mls1}. However, \textit{it is essential to note that FSL is typically a closed-set learning problem where the classes in the support and query sets are identical, and it may not account for extreme domain differences between training and testing.}\\
\noindent \textbf{Few-shot open-set learning}: FSL struggles to handle unknown class samples during testing. A common but sub-optimal approach is to use closed-set FSL models for outlier rejection by setting a confidence threshold. However, finding the optimal threshold is challenging. To address this issue, researchers have focused on dedicated FSOS techniques, such as forming Gaussian clusters from a few training samples \cite{liu2020few} or optimizing confidence for outlier detection \cite{jeong2021few}. However, these methods struggle with outliers that resemble known class samples. Recent advances in FSOS, including the binary closed-open discriminator optimized episodically \cite{pal2022few} and the use of generative augmentaton \cite{pal2023morgan}, have shown promising results in handling outliers. Nonetheless, these methods can not handle domain shifts between the training and test sets. \textit{Our augmentation strategy differs from both of \cite{pal2023morgan, pal2022few}, as we seek to generate pseudo-known and unknown samples of both the domains, as opposed to those single domain FSOS methods}.\\
\noindent \textbf{Domain adaptive few-shot learning:} In cross-domain few-shot learning, global or class-conditional domain shifts may occur between the disjoint base and novel classes. One straightforward approach is to combine an off-the-shelf FSL technique with a cross-domain alignment objective \cite{dann}. However, direct domain alignment \cite{berthelot2021adamatch} may not always be preferable due to the label shift between the source and target domains. To address this issue, \cite{zhao2021domain, cdfsl1} proposed a method to align the domains while preserving global class discrimination. Alternatively, \cite{cdfsl2} utilized cross-domain mix-up samples to learn generalized features. Lastly, \cite{cdfsl3} extended this setup beyond natural image classification by using a meta-learning-based approach. \textit{In contrast, we propose a simple and efficient batch-norm-based global cross-domain prototype alignment strategy while maintaining class separation through dedicated metric losses.}

\section{Proposed methodology}

\subsection{Problem definition and notations}
Under our DA-FSOS setting, we have access to a large training set $\mathcal{D}_s$ from a set of source classes $\mathcal{C}_s$ in a source domain $\mathcal{S}$, a few-shot sample set $\mathcal{D}_d$ from a set of target classes $\mathcal{C}_d$ in a target domain $\mathcal{T}$ ($P(\mathcal{S}) \neq P(\mathcal{T})$), and a test set $\mathcal{D}_t$ from another set of $\mathcal{C}_t$ classes in $\mathcal{T}$.
It is important to note that $\mathcal{C}_s \cap \mathcal{C}_d = \emptyset$, $\mathcal{C}_t \cap \mathcal{C}_d = \emptyset$, and $\mathcal{C}_s \cap \mathcal{C}_t = \emptyset$. Additionally, the classes in $\mathcal{C}_t$ consist of known and unknown classes, respectively. During testing, the few-shot support set $\mathcal{D}'_{t}$ contains samples for a subset of $\mathcal{C}'_t \subset \mathcal{C}_t$ known classes, while the remaining $\mathcal{C}_t - \mathcal{C}'_t$ classes in $\mathcal{D}_t$ are unknown during training.
Our goal is to use $\mathcal{D}_s$ and $\mathcal{D}_d$ to learn an open-set classifier that can distinguish the samples from $\mathcal{D}_t$ into one of the known classes or an unknown common class with respect to $\mathcal{D}'_t$. 

In our setting, each episode consists of a set of known and pseudo-unknown classes denoted by $\mathcal{K}_\mathcal{S}$ and $\mathcal{U}_\mathcal{S}$ from $\mathcal{S}$, where $\mathcal{K}_\mathcal{S}, \mathcal{U}_\mathcal{S} \in \mathcal{C}_s$. Similarly, from $\mathcal{T}$, the set of known and pseudo-unknown classes are represented as $\mathcal{K}_\mathcal{T}$ and $\mathcal{U}_\mathcal{T}$, where $\mathcal{K}_\mathcal{T}, \mathcal{U}_\mathcal{T} \in \mathcal{C}_d$. $\mathcal{K}_{\mathcal{S}} \neq \mathcal{U}_{\mathcal{S}}$ and $\mathcal{K}_{\mathcal{T}} \neq \mathcal{U}_{\mathcal{T}}$

To form the source support, we use $m_{\mathcal{S}}$ samples from each class in $\mathcal{K}_\mathcal{S}$, denoted by $\mathbb{S}_{\mathcal{S}} = \{(x_i^s,y_i^s)\}_{i=1}^{|\mathcal{K}_\mathcal{S}|* m_{\mathcal{S}} }$. Similarly, we use $m_{\mathcal{T}}$ samples from each class in $\mathcal{K}_\mathcal{T}$ to form the target support set, denoted by $\mathbb{S}_{\mathcal{T}} = \{(x_i^t,y_i^t)\}_{i=1}^{|\mathcal{K}_\mathcal{T}|* m_{\mathcal{T}}}$.
Our support set formation follows the $\mathcal{K}$-way $m$-shot learning protocol, where $m_{\mathcal{S}}+m_{\mathcal{T}}=m$, $|\mathcal{K}_\mathcal{S}|= |\mathcal{K}_\mathcal{T}| = |\mathcal{K}|$ and $|\mathcal{U}_\mathcal{S}|= |\mathcal{U}_\mathcal{T}| = |\mathcal{U}|$. 
Here, $|\mathcal{K}|$ represents the cardinality operator of $\mathcal{K}$.
Also, we employ more supervision in $\mathcal{S}$ and limited samples in $\mathcal{T}$ for episodic training of DA-FSOS. Thereby, $m_{\mathcal{S}}>m_{\mathcal{T}}$. We represent the query sets by $\mathbb{Q}_{\mathcal{S}}(\mathcal{K}_{\mathcal{S}} \cup   \mathcal{U}_{\mathcal{S}})$ and $\mathbb{Q}_{\mathcal{T}}(\mathcal{K}_{\mathcal{T}} \cup \mathcal{U}_{\mathcal{T}})$.

\subsection{Model and training overview: \textsc{DAFOS-Net}}

\noindent \textbf{Architecture overview:} Referring to Fig \ref{fig:Architecture}, \textsc{DAFOS-Net} is composed of several components. First, there is a pre-trained feature extractor, denoted as $f_{\varphi}$, that encodes the original images along with their augmented versions. In addition, there are two conditional GANs, represented as ($G_{\mathcal{L}\theta}, D_{\mathcal{L}\phi}$) and $(G_{\mathcal{H}\theta}, D_{\mathcal{H}\phi})$, each consisting of a generator and discriminator sub-network. These cGANs further augment the closed classes and the pseudo-open-space for both domains at the feature level, using two noise distributions, $\mathcal{N}(0,\sigma L)$ and $\mathcal{N}(0,\sigma H)$, where $\sigma H >> \sigma L$. The variance for the closed-set classes, $\sigma L$, is kept low to discourage the generation of sparse features. On the other hand, a high variance, $\sigma H$, is used to synthesize more diverse pseudo-open-space features.

Finally, we introduce two classifiers: a domain classification network denoted as $\mathcal{O}_{\eta}$, that predicts whether a sample belongs to the \textit{source} or \textit{target} domain, and an \textit{inlier/outlier} classification network, denoted as $\mathcal{O}_{\zeta}$, that aids in making better decisions. To ensure that the domains are aligned globally, we introduce the domain-specific batch-norm layer into \textsc{DAFOS-Net}. Specifically, we learn translation and scaling parameters for feature normalization on a per-batch basis for each domain, which are denoted as ($\gamma_{\mathcal{S}}, \beta_{\mathcal{S}}$) and ($\gamma_{\mathcal{T}}, \beta_{\mathcal{T}}$), respectively.

\noindent \textbf{Training overview:} 
Our training involves loss optimization based on samples from both domains. For 1-shot meta-training, we take each domain data in tandem in each episode. To begin with, we use $f_{\varphi}$ to generate feature embeddings for the support and query sets of both domains in $\mathbb{S}_{f\mathcal{S}/f\mathcal{T}}$ and $\mathbb{Q}_{f\mathcal{S}/f\mathcal{T}}$.
We train the cGANs for synthesizing domain-specific features according to the $\mathcal{K}_{\mathcal{S}}/\mathcal{K}_{\mathcal{T}}$ known classes and the $\mathcal{U}_{\mathcal{S}}/\mathcal{U}_{\mathcal{T}}$ pseudo-unknown classes, respectively (Eq. \ref{naive_GAN}). However, training cGANs from limited samples is challenging due to the lack of sufficient gradients to train the generators. To address this issue, we use DAWSON \cite{dawson}, an optimization-based meta-learning method, to bridge GANs' likelihood-invariant training and obtain gradients in the low-shot regime. Specifically, we use first-order optimization-based meta-learning \cite{nichol2018first}.

The discriminator $D_{\mathcal{L}\phi}$ penalizes the generator $G_{\mathcal{L}\theta}$ for producing slightly fake known features $\mathbb{S}_{l\mathcal{S}}/\mathbb{S}_{l\mathcal{T}}$, given the low noise variation $\sigma L$, such that the union of $\mathbb{S}_{f\mathcal{S}/f\mathcal{T}}$ and $\mathbb{S}_{l\mathcal{S}/ l\mathcal{T}}$ better estimates the density for the known classes of $\mathcal{S}$ and $\mathcal{T}$. Similarly, $D_{\mathcal{H}\phi}$ penalizes $G_{\mathcal{H}\theta}$, which considers a higher noise variance $\sigma H$, for producing highly fake pseudo-unknown features $\mathbb{S}_{h\mathcal{S}}/\mathbb{S}_{h\mathcal{T}}$. To avoid possible mode collapse for the samples from both the cGANs, we introduce a novel regularizer that penalizes the generation of similar features for identical noise vectors (Eq. \ref{AOL_eqn}).
Moving forward, we calculate the class prototypes $\mathcal{P}'_{\mathcal{S}}$ and $\mathcal{P}'_{\mathcal{T}}$ given $\mathbb{S}_{f\mathcal{S}} \cup \mathbb{S}_{l \mathcal{S}}$ and $\mathbb{S}_{f\mathcal{T}} \cup \mathbb{S}_{l \mathcal{T}}$, respectively (Eq. \ref{ProtoMeanEqn}). For domain alignment, we propose aligning the styles of $\mathcal{P}_{\mathcal{S}}$ and $\mathcal{P}_{\mathcal{T}}$ by normalizing the prototype features with respect to $(\gamma_{\mathcal{S}}, \beta_{\mathcal{S}})$ and $(\gamma_{\mathcal{T}}, \beta_{\mathcal{T}})$, and minimizing the global distance between all the normalized source and target prototypes (Eq. \ref{ProtoAlignEqn}).

To ensure the discriminativeness of the embedding space, we propose a supervised contrastive loss $\mathcal{L}_{C|\mathcal{S} \cup \mathcal{T}}$ (Eq. \ref{intra_loss}) to minimize the vectorized euclidean distance ($\mathcal{Q}_{dist}$) between a query sample and its respective prototype, which leads to more compactness, where, $\mathcal{Q}_{dist}= \{(\mathcal{Q}_{dist_{1}}, \cdots, \mathcal{Q}_{dist_{n}}), (\mathcal{Q}_{dist} = \Vert \mathbb{Q}_{f\mathcal{S}/f\mathcal{T}} - \mathcal{P}_{\mathcal{S}/\mathcal{T}}\Vert^2_{2}  )\}$. Additionally, we introduce a novel prototype diversification objective $\mathcal{L}_{PD}$ (Eq. \ref{Triplet_L}) that ensures the similarity between a prototype and its positive class samples is at least a margin $\alpha$ higher than the similarity between that prototype and all the class prototypes from the other domain and the in-domain pseudo-unknown class samples. Finally, we optimize the classification losses for $(\mathcal{O}_{\zeta}, \mathcal{O}_{\eta})$ given $\mathcal{Q}_{dist}$ (Eq. \ref{Outlierloss}-\ref{Domainloss}). In each episode, we train the model end-to-end concerning all the losses (Eq. \ref{TotalLoss}) and discuss them below.

\subsection{Loss functions, training, and inference}

\noindent \textbf{The cGAN training losses:} In Eq. \ref{naive_GAN}, we present the min-max loss objectives, $\mathcal{L}_h$ and $\mathcal{L}_l$, for the feature-generating cGANs that operate in pseudo-open and closed spaces. The real data for hallucinating closed space is  $\mathbb{S}_{f\mathcal{S}/f\mathcal{T}}(\mathcal{K}_{\mathcal{S}}/\mathcal{K}_{\mathcal{T}})$, and the same for the open space is $\mathbb{Q}_{f\mathcal{S}/f\mathcal{T}}(\mathcal{U}_{\mathcal{S}}/\mathcal{U}_{\mathcal{T}})$. The synthesized features from the closed and open space cGANs for both $\mathcal{S}$ and $\mathcal{T}$ are represented by $\mathbb{S}_{l\mathcal{S}/l\mathcal{T}}$ and $\mathbb{S}_{h\mathcal{S}/h\mathcal{T}}$, respectively. To ensure that the pseudo-open and closed feature synthesis does not overlap, we incorporate a novel regularizer (Eq. \ref{AOL_eqn}) in $\mathcal{L}_h$.
\begin{small}
\begin{align}\label{naive_GAN}
&\mathcal{L}_h = \underset{G_{\mathcal{H\theta}}}{min} \hspace{1mm} \underset{D_{\mathcal{H\phi}}}{max}\hspace{1mm} \mathbb{E}_{s \sim \mathbb{Q}_{f\mathcal{S}/f\mathcal{T}}(\mathcal{U}_{\mathcal{S}}/\mathcal{U}_{\mathcal{T}})} [log D_{\mathcal{H\phi}}(s|\mathcal{U}_{\mathcal{S}}/ \mathcal{U}_{\mathcal{T}})] \nonumber\\
     & + \mathbb{E}_{z_h \sim \mathcal{N}(0, \sigma H)}[log (1-D_{\mathcal{H\phi}} (G_{\mathcal{H\theta}}(z_h|\mathcal{U}_{\mathcal{S}}/ \mathcal{U}_{\mathcal{T}})] + \mathcal{L}_{AOCMC} \nonumber\\
&\mathcal{L}_l = \underset{G_{\mathcal{L\theta}}}{min} \hspace{1mm} \underset{D_{\mathcal{L\phi}}}{max}\hspace{1mm} \mathbb{E}_{s \sim \mathbb{S}_{fS/fT}} [log D_{\mathcal{L\phi}}(s|\mathcal{K}_{\mathcal{S}}/ \mathcal{K}_{\mathcal{T}}))] \nonumber \\
    &  + \mathbb{E}_{z_l \sim \mathcal{N}(0, \sigma L)}[log (1-D_{\mathcal{L\phi}} (G_{\mathcal{L\theta}}(z_l|\mathcal{K}_{\mathcal{S}}/ \mathcal{K}_{\mathcal{T}}))];
\end{align}
\end{small}

\noindent \textbf{Anti open close mode collapse loss:} To ensure a controlled generation of diverse open and closed space features for both the domains by $G_{\mathcal{H}\theta}$ and $G_{\mathcal{L}\theta}$, respectively, using inputs $z_l \in \mathcal{N}(0, \sigma L)$ and $z_h \in \mathcal{N}(0, \sigma H)$, it is essential to avoid generating identical $s_h \in \mathbb{S}_h$ and $s_l \in \mathbb{S}_l$ for similar $z_l$ and $z_h$ inputs to the cGANs. To this end, we propose the anti-open-close mode collapse loss ($\mathcal{L}_{AOCMC}$) during the optimization of the open-space cGAN parameters ($\mathcal{H}\theta, \mathcal{H}\phi$) in $\mathcal{L}_h$, which is as follows,
\begin{small}
\begin{flalign}\label{AOL_eqn}
\mathcal{L}_{AOCMC} = \underset{G_{\mathcal{H}\theta}, D_{\mathcal{H}\phi}}{\text{min}} {1+\log} \frac{1-( 
 \text{cos}(z_l, z_h))+\epsilon}{1-(\text{cos}(s_l,s_h))+\epsilon}
\end{flalign}
\end{small}
Ideally, the set of values for $z_l$ is a subset of those of $z_h$, which means that there is a high likelihood of generating the same mode for $s_h $ and $s_l $ when $(z_h, z_l)$ is sampled from the overlapping region of $\mathcal{N}(0, \sigma L) \cup \mathcal{N}(0, \sigma H)$, where $\text{cos}(z_l,z_h) \approx 1$. To prevent this, $\mathcal{L}_{AOCMC}$ is used to regulate the generation of such features by adjusting $G_{\mathcal{H}\theta}$ such that $\text{cos}(s_l,s_h)$ tends to zero instead. A small constant $\epsilon$ is used to handle numerical instability. Alternatively, in the mode collapse case,  $\cos(s_h,s_l) \rightarrow 1$ for $\cos(z_h,z_l) \rightarrow 1$ so the error remains large, and needs to be penalized.

\noindent \textbf{Class and domain conditional prototype computation:} To compute the known-class prototypes for the classes in $\mathcal{K}_{\mathcal{S}}$ and $\mathcal{K}_{\mathcal{T}}$, we use augmented support sets $\mathbb{S}_{f\mathcal{S}/f\mathcal{T}}\cup \mathbb{S}_{l\mathcal{S}/l\mathcal{T}}$ separately for both domains. Specifically, we compute the prototypes $\{\mathcal{P}_{\mathcal{S}/\mathcal{T}}^{'k} \}_{k=1}^{|\mathcal{K}_{\mathcal{S}}|/|\mathcal{K}_{\mathcal{T}}|}$ for each of the $k^{th}$ known classes, where $m_{\mathcal{S}/\mathcal{T}}^k$ is the number of support samples corresponding to the $k^{th}$ class for ($\mathcal{S}, \mathcal{T}$).
\begin{small}
\begin{equation}\label{ProtoMeanEqn}
\mathcal{P}_{\mathcal{S}}^{'k}= {\frac{1}{m_\mathcal{S}^k}}{\displaystyle\sum_{k=1}^{|\mathcal{K}_{\mathcal{S}}| }}{\mathbb{S}_{f\mathcal{S}}^k\cup \mathbb{S}_{l\mathcal{S}}^k};  \; \mathcal{P}_{\mathcal{T}}^{'k}= {\frac{1}{m_\mathcal{T}^k}}{\displaystyle\sum_{k=1}^{|\mathcal{K}_{\mathcal{T}}| }}{\mathbb{S}_{fT}^k\cup \mathbb{S}_{l\mathcal{T}}^k};
\end{equation}
\end{small}

\noindent \textbf{Global cross-domain prototype alignment loss:} To globally align $\mathcal{S}$ and $\mathcal{T}$, we propose to align the style information of both domains by learning domain-specific batch-norm (DSBN) parameters $(\gamma_{\mathcal{S}}, \beta_{\mathcal{S}})$ and $(\gamma_{\mathcal{T}}, \beta_{\mathcal{T}})$, respectively. Furthermore, given the rescaled source and target domain prototypes based on the batch-norm parameters, we seek to minimize the global average distance between all $\mathcal{S}$ and $\mathcal{T}$ prototype embeddings, which is expressed as,
\begin{footnotesize}
\begin{align}\label{ProtoAlignEqn}
\mathcal{L}_{Align}= {\frac{  {\displaystyle\sum_{k=1}^{|\mathcal{K}_{\mathcal{S}} |}} (\gamma_{\mathcal{S}} \times \mathcal{P}_{\mathcal{S}}^{'k} + \beta_{\mathcal{S}})  }{|\mathcal{K}_{\mathcal{S}}|}} - {\frac{  {\displaystyle\sum_{k=1}^{|\mathcal{K}_{\mathcal{T}} |}} (\gamma_{\mathcal{T}} \times \mathcal{P}_{\mathcal{T}}^{'k} + \beta_{\mathcal{T}})  }{|\mathcal{K}_{\mathcal{T}}|}} ;
\end{align}
\end{footnotesize}

Our objective is to achieve rapid domain alignment by aligning only the prototype vectors instead of the entire dataset. As the known-class compactness loss brings the query and support samples closer to the prototypes, aligning the prototypes implicitly aligns the domain distributions. In contrast, the original DSBN in \cite{chang2019domain} generates pseudo-labels from $\mathcal{T}$ and performs a two-stage approach for class-level domain alignment, which is suboptimal for DA-FSOS.

\noindent \textbf{Known-class compactness loss:} To maintain a concise and discriminative representation of each known class, minimizing the distance between the known support and query samples and the corresponding class prototypes is crucial. As described in Eq. \ref{intra_loss}, we enforce each $q \in \mathbb{Q}_{f\mathcal{S}}(\mathcal{K}_{\mathcal{S}})\cup\mathbb{S}_{f\mathcal{S}} \cup \mathbb{S}_{l\mathcal{S}}$ to be in close proximity to the respective prototype in $\mathcal{P}_{\mathcal{S}}$, and similarly for the target domain $\mathcal{T}$. The compactness loss for both domains is denoted as follows,
\begin{small}
\begin{equation}\label{intra_loss}
\mathcal{L}_{C|(\mathcal{S}\cup \mathcal{T})} = \mathbb{E}_{q \in \mathbb{Q}_{f}(\mathcal{K})\cup\mathbb{S}_{f} \cup \mathbb{S}_{l}} \left[  -\log\frac{e^ {-d(q,\mathcal{P}^l)}}{\sum_{k=1}^{\mathcal{K}}e^ {-d(q,\mathcal{P}^k)}}  \right]
\end{equation}
\end{small}
$d$ is the squared Euclidean distance and $\mathcal{P}^l$ is the class prototype for $q$. We ignore the domain labels in Eq. \ref{intra_loss} .

\noindent \textbf{Prototype diversification loss:} In \textsc{DAFOS-Net}, we strive to achieve not only within-category compactness, as ensured by Eq. \ref{intra_loss}, but also a margin of separation between open and closed spaces and class-discriminative feature space for both domains, which can be challenging to maintain due to domain adaptation. To address this, we propose a margin-based metric objective that enforces a significant distance between the prototypes in $\mathcal{P}_{\mathcal{S}}$ and the samples in $\mathbb{Q}_{f\mathcal{S}}(\mathcal{U}_\mathcal{S}) \cup \mathbb{S}_{h\mathcal{S}}$ from the same domain, as well as the cross-domain prototypes from $\mathcal{P}_{\mathcal{T}}$. This requirement applies vice versa for $\mathcal{T}$, resulting in highly discriminative feature space for both domains. 
\begin{small}
\begin{align}\label{Triplet_L}
&\mathcal{L}_{PD}=  \sum_{k=1}^{|\mathcal{K}_{\mathcal{S}}| }  {[ {\Vert \mathcal{P}_{\mathcal{S}}^{k}- {q}_{Pos|\mathcal{S}}\Vert_2^2} - {\Vert \mathcal{P}_{\mathcal{S}}^{k}- {q}_{Neg|\mathcal{S}}\Vert_2^2} + \alpha ]}_{+} \nonumber\\
&+ \sum_{k=1}^{|\mathcal{K}_{\mathcal{T}}| }  {[ {\Vert \mathcal{P}_{\mathcal{T}}^{k}- {q}_{Pos|\mathcal{T}}\Vert_2^2} - {\Vert \mathcal{P}_{\mathcal{T}}^{k}- {q}_{Neg|\mathcal{T}}\Vert_2^2} + \alpha ]}_{+};
\end{align}
\end{small}
Here, ${q}_{Pos|\mathcal{S}} \in \mathbb{Q}_{f\mathcal{S}} (\mathcal{K}_{\mathcal{S}} )^k$ refers to the positive queries from the source domain for the $k^{th}$ class prototype anchor $\mathcal{P}_{\mathcal{S}}^{k}$, while its negative queries are denoted by ${q}_{Neg|\mathcal{S}} \in \mathbb{Q}_{f\mathcal{S}}(\mathcal{U}_{\mathcal{S}} ) \cup \mathbb{S}_{h\mathcal{S}} \cup \mathcal{P}_{\mathcal{T}}$. Ideally, we seek to ensure that the difference between (anchor, +ve) should be less by a margin $\alpha$ than the difference between (anchor, -ve). Also, $ {[z]}_{+} = max(z, 0)$.

\noindent \textbf{Outlier detection and domain prediction loss:} We subsequently pass each query's distance $(\mathcal{Q}_{dist})$ from the prototypes to Outlier detector, $\mathcal{O}_{\zeta}$ and domain predictor $\mathcal{O}_{\eta}$. $\mathcal{O}_{\zeta}$ in Eq. \ref{Outlierloss} meta-learns to classify $\mathbb{Q}_{f\mathcal{S}}(\mathcal{K}_{\mathcal{S}}) \cup \mathbb{Q}_{f\mathcal{T}}(\mathcal{K}_{\mathcal{T}}) $ queries as known and query samples from $\mathbb{Q}_{f\mathcal{S}}(\mathcal{U}_{\mathcal{S}}) \cup \mathbb{Q}_{f\mathcal{T}}(\mathcal{U}_{\mathcal{T}}) \cup \mathbb{S}_{h\mathcal{S}}(\mathcal{U}_{\mathcal{S}}) \cup \mathbb{S}_{h\mathcal{T}}(\mathcal{U}_{\mathcal{T}})$ as outlier. 
\begin{small}
\begin{equation}\label{Outlierloss}
 \mathcal{L}_{OUT}  = \mathbb{E}_{q \in \{  \mathbb{Q}_{f\mathcal{S}}\cup \mathbb{Q}_{f\mathcal{T}}  \cup \mathbb{S}_{h\mathcal{S}}\cup \mathbb{S}_{h\mathcal{T}} \}}\left[-\sum_{i=1}^{2} t_{i}log(\mathcal{O}_{\zeta}(\mathcal{Q}_{dist})_{i})\right]
\end{equation}
\end{small}

Where, $t_{i} \in \{0,1\}$ represents \textit{known} or outlier label.

Similarly, $\mathcal{O}_{\eta}$ in Eq. \ref{Domainloss} learns to classify $\mathbb{Q}_{f\mathcal{S}}(\mathcal{K}_{\mathcal{S}}\cup\mathcal{U}_{\mathcal{S}}) \cup \mathbb{S}_{h\mathcal{S}}(\mathcal{U}_{\mathcal{S}})$ queries as source and $\mathbb{Q}_{f\mathcal{T}}(\mathcal{K}_{\mathcal{T}}\cup\mathcal{U}_{\mathcal{T}}) \cup \mathbb{S}_{h\mathcal{T}}(\mathcal{U}_{\mathcal{T}})$ queries as target.
\begin{small}
\begin{equation}\label{Domainloss}
 \mathcal{L}_{DC}  = \mathbb{E}_{q \in \{ \mathbb{Q}_{f\mathcal{S}}\cup \mathbb{S}_{h\mathcal{S}} \cup \mathbb{Q}_{f\mathcal{T}}  \cup \mathbb{S}_{h\mathcal{T}}  \}} \left[-\sum_{i=1}^{2} c_{i}log(\mathcal{O}_{\eta}(\mathcal{Q}_{dist})_{i})\right]
\end{equation}
\end{small}

Where, $c_{i} \in \{0,1\}$ indicates $\mathcal{S}$ or $\mathcal{T}$ domain label.

\noindent \textbf{Total loss:} To the end, we estimate the overall loss function to optimize $f_{\mathcal{\varphi}}$ in Eq. \ref{TotalLoss}. We follow an alternate optimization strategy between the cGANs and $f_{\varphi}$.
\begin{small}
\begin{equation}\label{TotalLoss}
\mathcal{L}_{FE} = \lambda_{1}\cdot\mathcal{L}_{C} + \lambda_{2}\cdot\mathcal{L}_{PD} + \lambda_{3}\cdot\mathcal{L}_{Align} + \lambda_{4}\cdot\mathcal{L}_{OUT} + \lambda_{5}\cdot\mathcal{L}_{DC}  
\end{equation}
\end{small}
Where, the $\lambda$s are weight factors of loss components.

\noindent \textbf{Inference}: In normal DA-FSOS, we utilize the outputs of $\mathcal{O}_{\zeta}$ to determine whether a sample is an inlier or outlier. For inlier samples, we estimate the class label based on their similarity with the prototype vectors calculated from $\mathcal{D}'_t$.

However, for generalized DA-FSOS, the query sample may come from any of the classes in $\mathcal{C}_s \cup \mathcal{C}_d \cup \mathcal{C}_t$. To handle this situation, we leverage both $\mathcal{O}_{\zeta}$ and $\mathcal{O}_{\eta}$ to determine both the domain and class labels for the test query. Subsequently, we choose the prototypes from the corresponding domain to classify the potential known-class queries.

\section{Experimental evaluations}\label{section:exp}
\noindent \textbf{Datasets, preprocessing, and network details}.
We evaluated the performance of \textsc{DAFOS-Net} on three benchmark datasets: Office-Home \cite{offhome}, DomainNet \cite{domainnet}, and mini-imageNet/CUB \cite{matchingnet,cub}. Office-Home contains 65 classes and four domains. For our experiments, we selected the \texttt{Real-World} and \texttt{Clipart} domains, and used the following class split: $\mathcal{C}_s = 25$, $\mathcal{C}_d = 10$, $\mathcal{C}'_t (\text{known}) = 15$, $\mathcal{C}_t - \mathcal{C}'_t(\text{unknown}) = 15$. The DomainNet dataset contains 345 classes and six domains. We considered the following source-target combinations: \texttt{Real} to \texttt{Clipart}, \texttt{Real} to \texttt{Painting}, and \texttt{Clipart} to \texttt{Painting}. The class split we used was $\mathcal{C}_s = 125$, $\mathcal{C}_d = 75$, $\mathcal{C}'_t (\text{known}) = 80$, $\mathcal{C}_t - \mathcal{C}'_t(\text{unknown}) = 65$. 
For the mini-imageNet and CUB datasets \cite{chen2019closer}, we used $\mathcal{C}_s = 100$, $\mathcal{C}_d = 50$, $\mathcal{C}'_t (\text{known}) = 25$, $\mathcal{C}_t - \mathcal{C}'_t(\text{unknown}) = 25$. Notably, CUB dataset has fine-grained separation among known / unknown classes. For all datasets, we resized the images to 84$\times$84 pixels. We consider two image-level augmentations, i) \textit{Weak} \cite{berthelot2021adamatch} (Flipping the images, rotation, resizing), and ii) \textit{Strong} \cite{cubuk2020randaugment}: intensity transformation with a random magnitude in $[0,0.5]$.

In \textsc{DAFOS-Net}, we utilize ResNet-18 \cite{resnet} for $f_{\varphi}$, and construct the generators $G_{\mathcal{H}\theta}$ and $G_{\mathcal{L}\theta}$ with four linear layers. Final batch norm layer in $f_{\varphi}$ is replaced by the DSBN layer. For all classifiers and cGAN discriminators, we use four linear layers followed by the respective output layers. In total, \textsc{DAFOS-Net} contains $5.8$M parameters, and training involves $0.08$ GFLOPS computations. (Fig \ref{fig:ablation1} (d)).

\noindent \textbf{Training and evaluation protocols}.
During meta-training, we use the Adam optimizer \cite{kingma2015adam} with a learning rate of 0.0001 and a batch size of 8. 
After cross validation, we set $\alpha=0.5$ in Eq. \ref{Triplet_L} (see \texttt{Supplementary}). The weights in Eq. \ref{TotalLoss} are cross-validated. For evaluation, we use the widely-used 5-way 1-shot and 5-way 5-shot training protocols consistent with previous works in DA-FSL and FSOS \cite{pal2023morgan, zhao2021domain}. We present the average top-1 accuracy for 500 test episodes to record  results over three independent runs.

To comprehensively evaluate \textsc{DAFOS-Net} performance, we utilize standard OSR metrics, which include closed-set classification accuracy (Acc) and Area Under ROC Curve (AUROC) \cite{bradley1997use}. Specifically, Acc measures the percentage of correctly classified known class samples, while AUROC evaluates model's ability to identify outliers.

\subsection{Comparison with the state-of-the-art}

\begin{table*}[!htbp]
\caption{The 5-way ($\mathcal{K}=5$) 1 and 5-shot DA-FSOS performance comparison of the proposed \textsc{DAFOS-Net} and SOTA Methods.}
\begin{center}
\scalebox{0.59}{
\begin{tabular}{|>{\columncolor{Gray}}c|>{\columncolor{Gray}}c|>{\columncolor{Gray}}c||c c ||c c ||c c | c c | c c|}
\hline
\rowcolor{Gray}
& & &\multicolumn{2}{c||}{\textbf{Office-Home}} &\multicolumn{2}{c||}{\textbf{MiniImageNet to CUB}}&\multicolumn{6}{c|}{\textbf{DomainNet}} \\
\rowcolor{Gray}
\textbf{Model}&\textbf{Venue}&\textbf{Paradigm}&\multicolumn{2}{c||}{\textbf{Real-World to Clipart}}
&\multicolumn{2}{c||}{}
&\multicolumn{2}{c}{\textbf{Real to Clipart}}&\multicolumn{2}{c}{\textbf{Real to Painting}}&\multicolumn{2}{c|}{\textbf{Clipart to Painting}}\\
\cline{4-13}
 \rowcolor{Gray}
 & & &\textbf{Acc(\%)}&\textbf{AUROC(\%)}&\textbf{Acc(\%)}&\textbf{AUROC(\%)}&\textbf{Acc(\%)}&\textbf{AUROC(\%)}&\textbf{Acc(\%)}&\textbf{AUROC(\%)}&\textbf{Acc(\%)}&\textbf{AUROC(\%)}\\
\hline
\hline
\rowcolor{Gray}
& & & \multicolumn{10}{c|}{\textbf{1-shot Evaluation}}\\
\cline{4-13}
PrototypicalNet \cite{protonet} & NIPS-17 & FSL &25.17 &19.23  &31.44  &25.12 &28.15 &23.02 &29.81 &23.51 &27.59 &22.18\\
Metaoptnet \cite{lee2019meta} & CVPR-19 & FSL &33.71 &26.62 &42.46  &34.22 &35.16 &27.19 &38.63 &26.02 &37.21  &29.56 \\
\hline
 OpenMax\cite{bendale2016towards} & CVPR-16 & OSR &12.19 &14.77 &21.88 &24.02 &14.68 &16.25  &15.71 &16.38 &15.66  &17.01 \\
\hline
PEELER \cite{liu2020few} & CVPR-20 & FSOS &14.55 &16.18 &25.47 &27.82 &18.81 &20.28 &21.06 &22.67 &22.64 &23.04 \\
SnaTCHer \cite{jeong2021few} & CVPR-21 & FSOS &20.36	&22.15	&31.33	&32.18	&23.79	&25.17	&24.43	&25.64	&25.18 &24.56\\
OCN \cite{pal2022few}  & WACV-22 & FSOS &22.64	&21.32	&35.98	&33.27	&24.11	&23.92	&26.24	&25.17	&27.65 &26.64 \\
MORGAN \cite{pal2023morgan} & WACV-23 & FSOS &33.92	&35.15	&40.62	&37.45	&33.15	&32.43	&34.89	&33.45	&39.22	&38.07 \\
\hline
AdaMatch \cite{berthelot2021adamatch} & ICLR-22 & DA &30.31	&27.29	&45.82	&36.26	&34.53	&30.36	&33.27	&29.71	&30.84	&26.34\\
\hline 
DAPN \cite{zhao2021domain} & WACV-21 & DA-FSL &31.86	&25.28	&47.82	&38.76	&38.44	&29.73	&41.65	&33.21	&43.49	&35.27 \\
NSAE \cite{liang2021boosting} & ICCV-21 & CDFSL &34.26	&28.17	&42.67	&34.59	&36.33	&28.25	&39.17	&31.07	&40.14	&33.19 \\
\hline
MORGAN + DAPN &	-&-  &36.22	&38.12	&41.12	&40.46	&36.58	&34.41	&38.75	&36.19	&41.77	&39.45 \\
\hline
\textsc{DAFOS-Net} [Ours] & - & DA-FSOS &\textbf{51.79$\pm$0.67}	&\textbf{50.54$\pm$0.54}	&\textbf{49.25$\pm$0.62}	&\textbf{50.17$\pm$0.28}	&\textbf{52.42$\pm$0.37}	&\textbf{54.84$\pm$0.36}	&\textbf{55.06$\pm$0.34}	&\textbf{54.16$\pm$0.33}	&\textbf{56.21$\pm$0.21}	&\textbf{57.26$\pm$0.49} \\
\hline
\hline

\rowcolor{Gray}
& & & \multicolumn{10}{c|}{\textbf{5-shot Evaluation}}\\
\cline{4-13}
PrototypicalNet \cite{protonet} & NIPS-17 & FSL &29.61	&24.75	&35.87	&31.27	&31.67	&23.43	&32.28	&25.96	&31.28	&24.98 \\
Metaoptnet \cite{lee2019meta} & CVPR-19 & FSL &36.51	&28.63	&46.36	&35.97	&38.45	&30.77	&41.73	&32.29	&40.42	&32.37 \\
\hline
 OpenMax\cite{bendale2016towards} & CVPR-16 & OSR &15.74	 &16.28	&26.43	&29.32	&16.26	&17.38	&17.46	&18.23	 &20.21	&19.45 \\
\hline
PEELER \cite{liu2020few} & CVPR-20 & FSOS  &20.16	&22.24	&30.71	&33.41	&22.52	&24.14	&23.24	&27.08	&24.82 	&25.12  \\
SnaTCHer \cite{jeong2021few} & CVPR-21 & FSOS &23.12	&25.12	&34.28	&35.13	&26.13	&27.16	&26.71	&28.16	&28.41 	&26.72\\
OCN \cite{pal2022few}  & WACV-22 & FSOS &24.28	&23.98	&37.42	&34.21	&25.86	&25.62	&27.87	&26.23	&31.35 &29.89  \\
MORGAN \cite{pal2023morgan} & WACV-23 & FSOS &35.31	&37.43	&42.04	&39.82	&36.25	&34.44	&38.32	&36.44	&42.05	&39.86 \\
\hline
AdaMatch \cite{berthelot2021adamatch} & ICLR-22 & DA &36.67	&30.48	&46.15	&39.19	&37.91	&31.21	&35.37	&31.34	&33.57	&28.38\\
\hline 
DAPN \cite{zhao2021domain} & WACV-21 & DA-FSL &34.55  &27.32	 &49.55	 &41.23	 &40.47	 &31.42	&44.29	&35.68	&45.57	&38.75 \\
NSAE \cite{liang2021boosting} & ICCV-21 & CDFSL &37.84	&29.65	&46.33	&36.67	&39.11	&29.32	&42.32	&34.42	&43.14	&35.23 \\
\hline
MORGAN + DAPN &-  &-  &39.52	&40.68	&45.55	&44.26	&40.64	&41.45	&40.27	&39.26	&45.81	&42.26\\
\hline
\textsc{DAFOS-Net} [Ours] & - & DA-FSOS &\textbf{54.44$\pm$0.29}	&\textbf{57.72$\pm$0.18}	&\textbf{58.51$\pm$0.32}	&\textbf{59.73$\pm$0.43}	&\textbf{55.09$\pm$0.64}	&\textbf{60.67$\pm$0.48}	&\textbf{59.03$\pm$0.25}	&\textbf{56.08$\pm$0.67}	&\textbf{59.66$\pm$0.28}	&\textbf{60.55$\pm$0.31} \\
\hline
\end{tabular}}
\label{tab_15shot}
\end{center}
\vspace*{-4mm}
\end{table*}

\renewcommand{\thefigure}{3}
\begin{figure*}[!htbp]
    \centering
    \includegraphics[width=\linewidth]{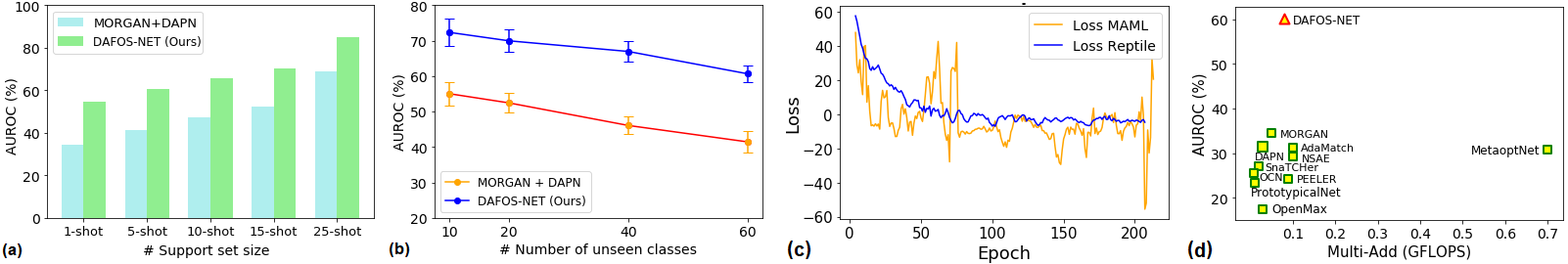}
    \vspace*{-1mm}
    \caption{AUROC variation due to a) support size, b) open classes for 5-way evaluation on DomainNet (\texttt{Real} to \texttt{clipart}) (c) Loss comparison due to optimizing \textsc{DAFOS-Net} by Reptile \cite{nichol2018first}, MAML \cite{finn2017model} d) Trade-off between  AUROC and GFLOPS by SOTA models.}
    \label{fig:ablation1}
    \vspace*{-3mm}
\end{figure*}

In Table \ref{tab_15shot}, we evaluate several state-of-the-art (SOTA) methodologies from various paradigms under the DA-FSOS settings and compare them against \textsc{DAFOS-Net}. 

We observe that PrototypicalNet \cite{protonet} and MetaoptNet \cite{lee2019meta}, which were originally developed for FSL, perform poorly in outlier recognition, especially under domain shift. As a result, their Acc and AUROC values are considerably worse than \textsc{DAFOS-Net}. While OpenMax \cite{bendale2016towards}, an OSR method, shows better outlier detection ability in AUROC than its Acc value, it performs poorly overall because its Weibull models are not effective in approximating distribution density from limited samples.

FSOS methods, including PEELER \cite{liu2020few}, SnaTCHer \cite{jeong2021few}, OCN \cite{pal2022few}, and MORGAN \cite{pal2023morgan}, are moderately effective in rejecting fine-grained outliers, but they lack transferable knowledge in $\mathcal{T}$. AdaMatch \cite{berthelot2021adamatch}, a domain adaptation method, and other DAFSL methods, namely, DAPN \cite{zhao2021domain} and NSAE \cite{liang2021boosting}, perform better in detecting known classes and generalize well over $\mathcal{T}$, but they fail to recognize fine-grained outliers in $\mathcal{T}$, for example in the CUB dataset consisting of different Birds categories.

Moreover, we incorporate \textit{Domain Adaptive Loss} from DAPN \cite{zhao2021domain} in optimizing MORGAN to adapt an FSOS framework over $\mathcal{T}$, and we observe better closed and open performance compared to their vanilla counterparts. We define this combined framework as MORGAN+DAPN. Nevertheless, overall, \textsc{DAFOS-Net} outperforms all other methodologies for both 1 and 5-shot settings due to its effective margin-aware outlier rejection and distribution alignment strategies. For instance, \textsc{DAFOS-Net} beats its next best alternative by $15.57\%$ Acc and a significant margin of $12.42\%$ AUROC for 1-shot evaluation over the Office-Home dataset. We observe similar dominant performance by \textsc{DAFOS-Net} for both 1 and 5-shot evaluations over Mini-imagenet to CUB and DomainNet datasets, reporting the new state-of-the-art. Finally, we evaluate the sensitivity of \textsc{DAFOS-Net} to the number of shots and the degree of openness and compare it against MORGAN+DAPN in Fig \ref{fig:ablation1} (a, b). The results demonstrate that \textsc{DAFOS-Net} outperforms MORGAN+DAPN in handling a larger set of open classes, with a consistent improvement of $15 \%$ AUROC.

\subsection{Generalized DA-FSOS comparison}

We conducted a thorough comparison of our proposed \textsc{DAFOS-Net} against MORGAN, DAPN, combination of MORGAN+DAPN for the generalized DA-FSOS setting. This setting is evaluated over all the source and target domain classes (refer to Table \ref{tab_generalized}). For all three datasets, \textsc{DAFOS-Net} outperforms MORGAN+DAPN by a significant margin. For instance, for DomainNet, \textsc{DAFOS-Net} has improvements of $14.69\%$ in closed-set accuracy and $21.79\%$ on the AUROC values over MORGAN+DAPN. Similar trends are observed for the other two datasets.

To assess the importance of the domain classifier $\mathcal{O}_{\eta}$ in classifying $\mathcal{S} \cup \mathcal{T}$ queries, we conducted further evaluations. We found that the model without $\mathcal{O}_{\eta}$ showed a significant drop in Acc and AUROC performances by almost $6\%$, averaging over both domains. This finding validates that the inclusion of $\mathcal{O}_{\eta}$  reduces the training bias of \textsc{DAFOS-Net}, while MORGAN+DAPN is much affected by overfitting.

\begin{table}[!htbp]
\caption{Generalized DA-FSOS comparison, where a query is sampled from $\mathcal{S}+\mathcal{T}$ in meta-testing ($\mathcal{K}=5, m=5$)}
\vspace*{-2mm}
\begin{center}
\scalebox{0.59}{
\begin{tabular}{|>{\columncolor{Gray}}c||c|c||c|c|}
\hline
\rowcolor{Gray}
\textbf{Dataset} &\textbf{Acc(\%)} &\textbf{AUROC(\%)} &\textbf{Acc(\%)} &\textbf{AUROC(\%)}  \\
\cline{2-5}
\rowcolor{Gray}
 & \multicolumn{2}{c||}{\textbf{MORGAN}}  &\multicolumn{2}{c|}{\textbf{\textsc{DAPN}}}\\
\hline

\hline
Office-Home (R $\rightarrow$ C)* &32.37$\pm$0.71    &30.65$\pm$0.09   &30.09$\pm$0.28  &27.16$\pm$0.67 \\ 
\hline
MiniImageNet to CUB &36.18$\pm$0.37    &34.89$\pm$0.82   &31.54$\pm$0.43   &26.08$\pm$0.58 \\ 
\hline
DomainNet (R $\rightarrow$ C)* &32.25$\pm$0.52    &35.53$\pm$0.71   &29.25$\pm$0.31   &28.67$\pm$0.45 \\ 
\hline
\rowcolor{Gray}
 & \multicolumn{2}{c||}{\textbf{MORGAN+DAPN}}  &\multicolumn{2}{c|}{\textbf{\textsc{DAFOS-Net}}}\\
\hline
Office-Home (R $\rightarrow$ C)* &36.23$\pm$0.17    &38.47$\pm$0.47   &\textbf{49.24$\pm$0.59}  &\textbf{52.25$\pm$0.83} \\ 
\hline
MiniImageNet to CUB &42.26$\pm$0.15    &41.06$\pm$0.31   &\textbf{52.79$\pm$0.62}   &\textbf{52.35$\pm$0.35} \\ 
\hline
DomainNet (R $\rightarrow$ C)* &38.23$\pm$0.06    &37.38$\pm$0.29   &\textbf{52.92$\pm$0.46}   &\textbf{59.17$\pm$0.18} \\ 
\hline
\multicolumn{5}{l}{*R: RealWorld domain, C: Clipart domain}
\end{tabular}}
\label{tab_generalized}
\end{center}
\vspace*{-8mm}
\end{table} 


\subsection{Ablation analysis}
\noindent \textbf{i. Ablation on the loss components}.
We conducted experiments to assess the individual contributions of the loss components in \textsc{DAFOS-Net} and report their performance in Table \ref{tab_loss}. Optimizing solely with $\mathcal{L}_{C}$, resulted in poor open-set performance since the model failed to distinguish outliers from known samples. Conversely, optimizing solely with $\mathcal{L}_{PD}$, leading to suboptimal Acc as the queries were not heavily pulled towards their respective prototype. However, combining $\mathcal{L}_{C}$ and $\mathcal{L}_{PD}$ resulted in improved Acc and AUROC compared to their individual presence.
\begin{table}
\caption{Performance comparison due to loss components for 5-way 5-shot evaluation on DomainNet (\texttt{Real} to \texttt{clipart}).}
\vspace*{-2mm}
\begin{center}
\scalebox{0.59}{
\begin{tabular}{|>{\columncolor{Gray}}c|c|c|}
\hline
\rowcolor{Gray}
\textbf{Loss function} &\textbf{Acc(\%)} &\textbf{AUROC(\%)} \\
\hline
\hline
$\mathcal{L}_{C}$ &45.38$\pm$0.56   & 41.64$\pm$0.38   \\
$\mathcal{L}_{PD}$ &  43.48$\pm$0.47   & 52.09$\pm$0.35 \\
 $\mathcal{L}_{C}+\mathcal{L}_{PD}$ & 47.76$\pm$0.71   & 56.27$\pm$0.59 \\
 $\mathcal{L}_{C}+\mathcal{L}_{PD}+\mathcal{L}_{Align}$ & 52.46$\pm$0.48   & 58.29$\pm$0.31 \\
$\mathcal{L}_{C}+\mathcal{L}_{PD}+\mathcal{L}_{Align}+\mathcal{L}_{OUT}$  & 53.79$\pm$0.43  & 59.38$\pm$0.35  \\
\hline
\hline
Without $\mathcal{L}_{AOCMC}$ regularizer  & 54.32$\pm$0.49   & 58.45$\pm$0.36 \\
\hline
\hline
Intermediate DSBN layers  & 51.56$\pm$0.29 & 53.09$\pm$0.64\\
\hline
\hline
$\mathcal{L}_{C}+\mathcal{L}_{PD}+\mathcal{L}_{Align}+\mathcal{L}_{OUT}+\mathcal{L}_{DC}$ + $\mathcal{L}_{AOCMC}$  &\textbf{55.09$\pm$0.64}   &\textbf{60.67$\pm$0.48}\\
\hline
\end{tabular}}
\label{tab_loss}
\end{center}
\vspace*{-8mm}
\end{table}
$\mathcal{L}_{Align}$, significantly improved Acc by aligning $\mathcal{S}$ and $\mathcal{T}$, enabling the model to estimate $\mathcal{T}$ domain known samples better following the $\mathcal{S}$ domain compaction. Furthermore, the addition of $\mathcal{L}_{OUT}$, enhanced AUROC performance due to its ability to distinguish pseudo-unknown queries. We have already discussed the importance of $\mathcal{L}_{DC}$ for generalized DA-FSOS. Finally, regularizing with $\mathcal{L}_{AOCMC}$ helped generate discriminative adversarial open samples, thereby boosting AUROC by almost $2 \%$. Applying DSBN at the final layer of $f_{\varphi}$ instead of all intermediate layers (at all batch norm positions of ResNet-18) helps to boost performance due to global domain alignment with discriminative features.

\noindent \textbf{ii.Optimization stability of cGANs.}
\textsc{DAFOS-Net} leverages a dual-GAN approach to generate adversarial known and outlier samples concurrently. Second-order optimization based on MAML \cite{finn2017model} for this dual-GAN results in oscillations and longer convergence time compared to first-order optimization based on Reptile \cite{nichol2018first}, as demonstrated in Fig. \ref{fig:ablation1} (c). The first-order optimization method promotes training stability while dealing with limited training samples.

\begin{table}[!htbp]
\caption{Performanace comparison due to different augmentations for 5-way 5-shot evaluation on DomainNet (\texttt{Real} to \texttt{clipArt})}
\vspace*{-3mm}
\begin{center}
\scalebox{0.59}{
\begin{tabular}{|>{\columncolor{Gray}}c|c|c|}
\hline
\rowcolor{Gray}
\textbf{Augmentation type} &\textbf{Acc(\%)} &\textbf{AUROC(\%)}  \\
\hline
\hline
W/o augmentation & 50.32$\pm$0.48  &43.06$\pm$0.41 \\
\hline
Image level augmentation & 51.88$\pm$0.76   &48.62$\pm$0.31 \\
\hline
Image + Feature level augmentation &\textbf{55.09$\pm$0.64}   &\textbf{60.67$\pm$0.48} \\
\hline
\end{tabular}}
\label{tab_Augmentation}
\end{center}
\vspace*{-5mm}
\end{table}

\begin{table}[!htbp]
\caption{Analysis of noise variance $(\sigma L, \sigma H)$ values on $G_{\mathcal{L\theta}}, G_{\mathcal{H\theta}}$ over DomainNet (\texttt{Real} to \texttt{clipArt}) [5-way 5-shot]}
\vspace*{-3mm}
\begin{center}
\scalebox{0.59}{
\begin{tabular}{|>{\columncolor{Gray}}c|>{\columncolor{Gray}}c|c|c|c|}
\hline
\rowcolor{Gray}
$\sigma L$ & $\sigma H$ &\textbf{Acc(\%)} &\textbf{AUROC(\%)}  &\textbf{KL divergence} \\
\hline
\hline
0.4 & 0.8 & 56.05$\pm$0.16  & 57.38$\pm$0.29 & 0.57  \\
\hline
0.5 & 0.8 & 56.11$\pm$0.51 & 56.43$\pm$0.51 & 0.59\\
0.7 & 0.8 & 57.02$\pm$0.48 & 54.15$\pm$0.23 & 0.58\\
\hline
0.3 & 0.9 &\textbf{55.09$\pm$0.64}   &\textbf{60.67$\pm$0.48} & \textbf{0.45}\\
0.1 & 1.0 & 53.37$\pm$0.68 & 62.61$\pm$0.17 & 0.31 \\
\hline
0.4 & 1.2 & 58.68$\pm$0.24 & 55.54$\pm$0.34 & 0.58 \\
\hline
\end{tabular}}
\label{tab_noise}
\end{center}
\vspace*{-7mm}
\end{table}

\noindent \textbf{iii. Importance of the proposed augmentations:}
In Table \ref{tab_Augmentation}, we examine the impact of image space enhancement using weak \cite{berthelot2021adamatch} and strong augmentations \cite{cubuk2020randaugment} and find that it results in a notable $5.6\%$ improvement in AUROC. Additionally, enhancing data density through dual-GAN generated adversarial closed-open features leads to a boost of $4.78\%$ in Acc and $17.61\%$ in AUROC compared to the model without any augmentation. Jointly maximizing image and feature space augmentation schemes results in a better approximation of the known class distribution, thereby minimizing the few-shot risk.


\noindent \textbf{iv. Sensitivity  to $(\sigma H, \sigma L)$:}
Table \ref{tab_noise} presents the impact of varying the standard deviations of two noise vectors on the quality of samples generated by both the cGANs. To measure the similarity between adversarial known samples and their real counterparts, we report the Kullback-Leibler (KL) divergence. Samples with slightly higher KL divergence \cite{hershey2007approximating} produce greater variety and enhance the density of the known classes.
In this regard, we observed three cases, which are also illustrated in the \texttt{Supplementary}: a) increasing $\sigma L$ and keeping $\sigma H$ constant resulted in higher Acc and KL divergence, as many discriminative adversarial known features were generated. However, AUROC decreased due to some limited pseudo-unknown queries being misclassified as known samples. b) decreasing $\sigma L$ and increasing $\sigma H$ reduced Acc and KL divergence, as adversarial known samples were generated with a distribution similar to that of real known samples. AUROC increased due to an enriched open space. c) extending $\sigma H$ beyond 1 with constant $\sigma L$ created a potential risk of mixing pseudo-unknown samples with other known class samples, resulting in false positives for Acc.
The best performance was achieved by combining $\sigma L = 0.3$ and $\sigma H = 0.9$.

\noindent \textbf{v. Comparison among domain adaptation losses:} We compared three methods, \textit{Domain Adaptive Loss} \cite{zhao2021domain}, vanilla \textit{Batch Normalization} \cite{ioffe2015batch} and \textit{Domain-Specific Batch Normalization (DSBN)} in \cite{chang2019domain} against our GCDPA loss. 
\begin{table}[!htbp]
\caption{Analysis of different domain adaptation losses on DomainNet (\texttt{Real} to \texttt{clipArt}) for 5-way 5-shot evaluation.}
\begin{center}
\vspace*{-2mm}
\scalebox{0.59}{
\begin{tabular}{|>{\columncolor{Gray}}c|c|c|c|}
\hline
\rowcolor{Gray}
\textbf{Adaptation Method} &\textbf{Acc(\%)} &\textbf{AUROC(\%)} & \textbf{Convergence} \\
\hline
\hline
Domain Adaptive Loss \cite{zhao2021domain} & 49.37$\pm$0.23 & 50.28$\pm$0.35 & 450 epochs\\
Batch Normalization \cite{ioffe2015batch}  & 49.02$\pm$0.33 & 48.45$\pm$0.23 & 265 epochs\\
Domain-Specific Batch Norm \cite{chang2019domain} & 48.18$\pm$0.36 & 46.76$\pm$0.51 & 288 epochs\\
\hline
Global Cross-Domain Prototype Alignment  &\textbf{55.09$\pm$0.64}   &\textbf{60.67$\pm$0.48} & \textbf{180 epochs}\\
\hline
\end{tabular}}
\label{tab_adaptation}
\end{center}
\vspace*{-3mm}
\end{table}
\renewcommand{\thefigure}{4}
\begin{figure}[!htbp]
    \centering
    \includegraphics[width=\linewidth]{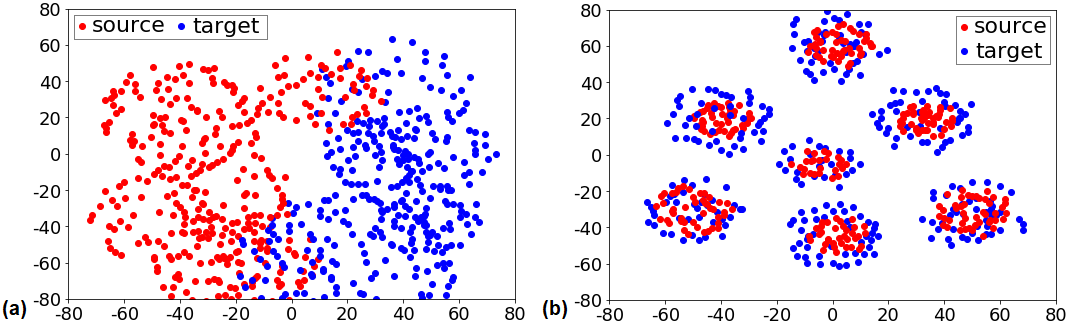}
    \caption{t-SNE visualization a) without and b) with adaptation.}
    \label{fig:tsne}
    \vspace*{-4mm}
\end{figure}
Distribution alignment through \cite{zhao2021domain} enhances model robustness, but batch-norm \cite{ioffe2015batch} helps to converge faster. However, due to globally aligning domains through prototypes by \textit{GCDPA} loss, \textsc{DAFOS-Net} converges faster and generalizes better than DSBN-based variant \cite{chang2019domain}, which aligns domains based on individual query samples.  
As per Table \ref{tab_adaptation}, the distribution alignment \cite{zhao2021domain} resulted in a $5.72\%$ lower Acc and $10\%$ lower AUROC compared to our \textit{GCDPA}-based adaptation. Time analysis also indicates that our approach achieves convergence quicker than \cite{zhao2021domain,chang2019domain}.

\noindent \textbf{vi. t-SNE visualization of metric space.}
The t-SNE \cite{tsne} visualization in Fig \ref{fig:tsne} illustrates the learning of \textsc{DAFOS-Net} with and without adaptation. For visual clarity, we presented seven categories from the \texttt{Real} and \texttt{Clipart} domains of DomainNet: Axe, Bus, Bucket, Clock, Flower, Foot, and Strawberry. The visualization reveals that the categories in $\mathcal{S}$ form compact clusters due to optimizing by $\mathcal{L}_{C}$. Furthermore, thanks to $\mathcal{L}_{Align}$, $\mathcal{T}$ categories maintain comparable semantic relationships while reducing domain shifts for each known class. For illustration simplicity, we did not evaluate the impact of $\mathcal{L}_{PD}$ in Fig \ref{fig:tsne} since we are considering the same set of classes from $\mathcal{S}$ and $\mathcal{T}$.

\section{Takeaways}
In this paper, we introduce a new problem setting for domain adaptive few-shot open-set learning and propose a novel architecture called \textsc{DAFOS-Net} to address it. Our approach ensures the modeling of a discriminative and domain-invariant embedding space while being able to reject outliers during testing. To achieve this, we propose a generative augmentation strategy for both the known classes and pseudo-open space for both domains, a domain-centric batch-norm parameter learning-based GCDPA loss, and a prototype diversification objective to ensure discriminativeness, in our meta-learning algorithm. We evaluate our proposed method on benchmark datasets from the domain adaptation literature in both the standard and generalized DA-FSOS settings. Our results demonstrate that \textsc{DAFOS-Net} significantly outperforms all competitors in both settings. We hope that our research can be applied to various domains such as edge AI, medical imaging, remote sensing, and other safety-critical areas that deal with domain shifts, low-shot learning, and open-set learning in a joint fashion.

{\small
\bibliographystyle{ieee_fullname}
\bibliography{main}

\begin{thebibliography}{10}\itemsep=-1pt

\bibitem{bendale2016towards}
Abhijit Bendale and Terrance~E Boult.
\newblock Towards open set deep networks.
\newblock In {\em Proceedings of the IEEE conference on computer vision and pattern recognition}, pages 1563--1572, 2016.

\bibitem{berthelot2021adamatch}
David Berthelot, Rebecca Roelofs, Kihyuk Sohn, Nicholas Carlini, and Alex Kurakin.
\newblock Adamatch: A unified approach to semi-supervised learning and domain adaptation.
\newblock {\em arXiv preprint arXiv:2106.04732}, 2021.

\bibitem{bradley1997use}
Andrew~P Bradley.
\newblock The use of the area under the roc curve in the evaluation of machine learning algorithms.
\newblock {\em Pattern recognition}, 30(7):1145--1159, 1997.

\bibitem{chang2019domain}
Woong-Gi Chang, Tackgeun You, Seonguk Seo, Suha Kwak, and Bohyung Han.
\newblock Domain-specific batch normalization for unsupervised domain adaptation.
\newblock In {\em Proceedings of the IEEE/CVF conference on Computer Vision and Pattern Recognition}, pages 7354--7362, 2019.

\bibitem{fslaug2}
Xuewei Chao and Lixin Zhang.
\newblock Few-shot imbalanced classification based on data augmentation.
\newblock {\em Multimedia Systems}, pages 1--9, 2021.

\bibitem{ssl2}
Da Chen, Yuefeng Chen, Yuhong Li, Feng Mao, Yuan He, and Hui Xue.
\newblock Self-supervised learning for few-shot image classification.
\newblock In {\em ICASSP 2021-2021 IEEE International Conference on Acoustics, Speech and Signal Processing (ICASSP)}, pages 1745--1749. IEEE, 2021.

\bibitem{chen2019closer}
Wei-Yu Chen, Yen-Cheng Liu, Zsolt Kira, Yu-Chiang~Frank Wang, and Jia-Bin Huang.
\newblock A closer look at few-shot classification.
\newblock {\em arXiv preprint arXiv:1904.04232}, 2019.

\bibitem{cubuk2020randaugment}
Ekin~D Cubuk, Barret Zoph, Jonathon Shlens, and Quoc~V Le.
\newblock Randaugment: Practical automated data augmentation with a reduced search space.
\newblock In {\em Proceedings of the IEEE/CVF conference on computer vision and pattern recognition workshops}, pages 702--703, 2020.

\bibitem{finn2017model}
Chelsea Finn, Pieter Abbeel, and Sergey Levine.
\newblock Model-agnostic meta-learning for fast adaptation of deep networks.
\newblock In {\em International Conference on Machine Learning}, pages 1126--1135. PMLR, 2017.

\bibitem{cdfsl2}
Yuqian Fu, Yanwei Fu, and Yu-Gang Jiang.
\newblock Meta-fdmixup: Cross-domain few-shot learning guided by labeled target data.
\newblock In {\em Proceedings of the 29th ACM International Conference on Multimedia}, pages 5326--5334, 2021.

\bibitem{dann}
Yaroslav Ganin and Victor Lempitsky.
\newblock Unsupervised domain adaptation by backpropagation.
\newblock In {\em International conference on machine learning}, pages 1180--1189. PMLR, 2015.

\bibitem{osrsurvey}
Chuanxing Geng, Sheng-jun Huang, and Songcan Chen.
\newblock Recent advances in open set recognition: A survey.
\newblock {\em IEEE transactions on pattern analysis and machine intelligence}, 43(10):3614--3631, 2020.

\bibitem{cdfsl3}
Yunhui Guo, Noel~C Codella, Leonid Karlinsky, James~V Codella, John~R Smith, Kate Saenko, Tajana Rosing, and Rogerio Feris.
\newblock A broader study of cross-domain few-shot learning.
\newblock In {\em Computer Vision--ECCV 2020: 16th European Conference, Glasgow, UK, August 23--28, 2020, Proceedings, Part XXVII 16}, pages 124--141. Springer, 2020.

\bibitem{resnet}
Kaiming He, Xiangyu Zhang, Shaoqing Ren, and Jian Sun.
\newblock Deep residual learning for image recognition.
\newblock In {\em Proceedings of the IEEE conference on computer vision and pattern recognition}, pages 770--778, 2016.

\bibitem{hershey2007approximating}
John~R Hershey and Peder~A Olsen.
\newblock Approximating the kullback leibler divergence between gaussian mixture models.
\newblock In {\em 2007 IEEE International Conference on Acoustics, Speech and Signal Processing-ICASSP'07}, volume~4, pages IV--317. IEEE, 2007.

\bibitem{ioffe2015batch}
Sergey Ioffe and Christian Szegedy.
\newblock Batch normalization: Accelerating deep network training by reducing internal covariate shift.
\newblock In {\em International conference on machine learning}, pages 448--456. pmlr, 2015.

\bibitem{jeong2021few}
Minki Jeong, Seokeon Choi, and Changick Kim.
\newblock Few-shot open-set recognition by transformation consistency.
\newblock In {\em Proceedings of the IEEE/CVF Conference on Computer Vision and Pattern Recognition}, pages 12566--12575, 2021.

\bibitem{ssl1}
Zilong Ji, Xiaolong Zou, Tiejun Huang, and Si Wu.
\newblock Unsupervised few-shot learning via self-supervised training.
\newblock {\em arXiv preprint arXiv:1912.12178}, 2019.

\bibitem{relationalnet}
Dahyun Kang, Heeseung Kwon, Juhong Min, and Minsu Cho.
\newblock Relational embedding for few-shot classification.
\newblock In {\em Proceedings of the IEEE/CVF International Conference on Computer Vision}, pages 8822--8833, 2021.

\bibitem{kingma2015adam}
Diederik~P Kingma and Jimmy Ba.
\newblock Adam: A method for stochastic optimization. iclr. 2015.
\newblock {\em arXiv preprint arXiv:1412.6980}, 9, 2015.

\bibitem{alexnet}
Alex Krizhevsky, Ilya Sutskever, and Geoffrey~E Hinton.
\newblock Imagenet classification with deep convolutional neural networks.
\newblock {\em Communications of the ACM}, 60(6):84--90, 2017.

\bibitem{lee2019meta}
Kwonjoon Lee, Subhransu Maji, Avinash Ravichandran, and Stefano Soatto.
\newblock Meta-learning with differentiable convex optimization.
\newblock In {\em Proceedings of the IEEE/CVF conference on computer vision and pattern recognition}, pages 10657--10665, 2019.

\bibitem{liang2021boosting}
Hanwen Liang, Qiong Zhang, Peng Dai, and Juwei Lu.
\newblock Boosting the generalization capability in cross-domain few-shot learning via noise-enhanced supervised autoencoder.
\newblock In {\em Proceedings of the IEEE/CVF International Conference on Computer Vision}, pages 9424--9434, 2021.

\bibitem{dawson}
Weixin Liang, Zixuan Liu, and Can Liu.
\newblock Dawson: A domain adaptive few shot generation framework.
\newblock {\em arXiv preprint arXiv:2001.00576}, 2020.

\bibitem{liu2020few}
Bo Liu, Hao Kang, Haoxiang Li, Gang Hua, and Nuno Vasconcelos.
\newblock Few-shot open-set recognition using meta-learning.
\newblock In {\em Proceedings of the IEEE/CVF Conference on Computer Vision and Pattern Recognition}, pages 8798--8807, 2020.

\bibitem{nichol2018first}
Alex Nichol, Joshua Achiam, and John Schulman.
\newblock On first-order meta-learning algorithms.
\newblock {\em arXiv preprint arXiv:1803.02999}, 2018.

\bibitem{pal2023morgan}
Debabrata Pal, Shirsha Bose, Biplab Banerjee, and Yogananda Jeppu.
\newblock Morgan: Meta-learning-based few-shot open-set recognition via generative adversarial network.
\newblock In {\em Proceedings of the IEEE/CVF Winter Conference on Applications of Computer Vision}, pages 6295--6304, 2023.

\bibitem{pal2022few}
Debabrata Pal, Valay Bundele, Renuka Sharma, Biplab Banerjee, and Yogananda Jeppu.
\newblock Few-shot open-set recognition of hyperspectral images with outlier calibration network.
\newblock In {\em Proceedings of the IEEE/CVF Winter Conference on Applications of Computer Vision}, pages 3801--3810, 2022.

\bibitem{osda}
Pau Panareda~Busto and Juergen Gall.
\newblock Open set domain adaptation.
\newblock In {\em Proceedings of the IEEE international conference on computer vision}, pages 754--763, 2017.

\bibitem{fsls2}
Archit Parnami and Minwoo Lee.
\newblock Learning from few examples: A summary of approaches to few-shot learning.
\newblock {\em arXiv preprint arXiv:2203.04291}, 2022.

\bibitem{domainnet}
Xingchao Peng, Qinxun Bai, Xide Xia, Zijun Huang, Kate Saenko, and Bo Wang.
\newblock Moment matching for multi-source domain adaptation.
\newblock In {\em Proceedings of the IEEE/CVF international conference on computer vision}, pages 1406--1415, 2019.

\bibitem{mlssl}
Mengye Ren, Eleni Triantafillou, Sachin Ravi, Jake Snell, Kevin Swersky, Joshua~B Tenenbaum, Hugo Larochelle, and Richard~S Zemel.
\newblock Meta-learning for semi-supervised few-shot classification.
\newblock {\em arXiv preprint arXiv:1803.00676}, 2018.

\bibitem{protonet}
Jake Snell, Kevin Swersky, and Richard Zemel.
\newblock Prototypical networks for few-shot learning.
\newblock {\em Advances in neural information processing systems}, 30, 2017.

\bibitem{cdfsl1}
Hung-Yu Tseng, Hsin-Ying Lee, Jia-Bin Huang, and Ming-Hsuan Yang.
\newblock Cross-domain few-shot classification via learned feature-wise transformation.
\newblock {\em arXiv preprint arXiv:2001.08735}, 2020.

\bibitem{tsne}
Laurens Van~der Maaten and Geoffrey Hinton.
\newblock Visualizing data using t-sne.
\newblock {\em Journal of machine learning research}, 9(11), 2008.

\bibitem{mls1}
Joaquin Vanschoren.
\newblock Meta-learning: A survey.
\newblock {\em arXiv preprint arXiv:1810.03548}, 2018.

\bibitem{offhome}
Hemanth Venkateswara, Jose Eusebio, Shayok Chakraborty, and Sethuraman Panchanathan.
\newblock Deep hashing network for unsupervised domain adaptation.
\newblock In {\em Proceedings of the IEEE conference on computer vision and pattern recognition}, pages 5018--5027, 2017.

\bibitem{matchingnet}
Oriol Vinyals, Charles Blundell, Timothy Lillicrap, Daan Wierstra, et~al.
\newblock Matching networks for one shot learning.
\newblock {\em Advances in neural information processing systems}, 29, 2016.

\bibitem{cub}
Catherine Wah, Steve Branson, Peter Welinder, Pietro Perona, and Serge Belongie.
\newblock The caltech-ucsd birds-200-2011 dataset.
\newblock 2011.

\bibitem{wang2020generalizing}
Yaqing Wang, Quanming Yao, James~T Kwok, and Lionel~M Ni.
\newblock Generalizing from a few examples: A survey on few-shot learning.
\newblock {\em ACM computing surveys (csur)}, 53(3):1--34, 2020.

\bibitem{zhao2021domain}
An Zhao, Mingyu Ding, Zhiwu Lu, Tao Xiang, Yulei Niu, Jiechao Guan, and Ji-Rong Wen.
\newblock Domain-adaptive few-shot learning.
\newblock In {\em Proceedings of the IEEE/CVF Winter Conference on Applications of Computer Vision}, pages 1390--1399, 2021.

\bibitem{fslaug1}
Jing Zhou, Yanan Zheng, Jie Tang, Jian Li, and Zhilin Yang.
\newblock Flipda: Effective and robust data augmentation for few-shot learning.
\newblock {\em arXiv preprint arXiv:2108.06332}, 2021.

\end{thebibliography}
}

\end{document}